\documentclass[10pt,twocolumn,letterpaper]{article}

\usepackage[
    top=0.75in,
    bottom=0.75in,
    left=0.75in,
    right=0.75in
]{geometry}
\usepackage[T1]{fontenc}
\usepackage{amsmath,amssymb}
\usepackage{newtxtext,newtxmath}
\usepackage{microtype}

\usepackage{graphicx}
\usepackage{caption}
\usepackage{booktabs}
\usepackage{multirow}
\usepackage{xcolor}

\usepackage[numbers,square,sort&compress]{natbib}
\usepackage[hyphens]{url}

\usepackage{algorithm}
\usepackage{algorithmic}

\usepackage{newfloat}
\usepackage{listings}

\DeclareCaptionStyle{ruled}{
    labelfont=normalfont,
    labelsep=colon,
    strut=off
}

\floatstyle{ruled}
\newfloat{listing}{tb}{lst}{}
\floatname{listing}{Listing}

\usepackage{titling}

\date{}

\usepackage[colorlinks=true,allcolors=blue]{hyperref}

\usepackage{multirow,color,colortbl,soul,bm,booktabs,amssymb,amsmath,amsfonts,xspace,caption}
\usepackage{algorithm}
\usepackage{algorithmic}
\usepackage{tabularx}
\usepackage[misc]{ifsym}

\newcommand{\boldstart}[1]{\noindent\textbf{#1}}

\usepackage{etoolbox}
\newtoggle{withSupp}
\toggletrue{withSupp}

\newcommand{\ours}{BudgetSplat\xspace}

\title{
    BudgetSplat: Budget-Controllable Generalizable Gaussian Splatting
}

\author{
    Yuxi Hu\textsuperscript{1},
    Jun Zhang\textsuperscript{1},
    Wei Mao,
    Yuchen Rao\textsuperscript{1},
    Friedrich Fraundorfer\textsuperscript{1}
    \\[0.6em]
    \textsuperscript{1}Graz University of Technology
     \\[0.3em]
    \textsuperscript{1}\textnormal{firstname.lastname@tugraz.at}\\
    \textsuperscript{2}\textnormal{wei.mao.research@gmail.com}
}

\date{}

\begin{document}

\maketitle

% -------------------------------------------------------
%%%%%%%%% ABSTRACT START
\begin{abstract}
Feed-forward 3D Gaussian Splatting enables efficient generalizable novel view synthesis, but existing pixel-aligned methods often generate highly redundant Gaussian primitives, causing the final representation size and rendering cost to grow rapidly with image resolution and the
number of input views. We propose BudgetSplat, a budget-controllable feed-forward framework that adapts the final Gaussian representation to a user-specified preservation ratio. Rather than treating compaction as simple importance ranking, we formulate it as a joint problem of importance estimation, spatial coverage, and cross-view redundancy. A single trained model supports varying numbers of context views and multiple test-time budgets without per-budget retraining or per-scene compression. To realize this formulation, Intra-Cross Mask Supervision provides structure-, coverage-, and redundancy-aware guidance for learning Gaussian importance, while budget-conditioned feature modulation adapts Gaussian opacity and attributes to the requested preservation ratio. Coverage-Aware Budget Selection further combines patch-wise pre-selection, view-wise opacity ranking, and 3D redundancy suppression to construct compact yet spatially distributed Gaussian representations. Experiments on indoor reconstruction and cross-dataset generalization demonstrate consistent quality--size trade-offs across different input view counts and preservation ratios, substantially reducing final representation storage and rendering cost while preserving rendering quality.
\end{abstract}

%%%%%%%%% ABSTRACT END

%%%%%%%%% BODY TEXT START
\section{1. Introduction}
\label{sec:intro}
3D Gaussian Splatting (3D-GS)~\cite{3dgs} represents scenes using explicit anisotropic Gaussians and enables high-quality real-time novel view synthesis. However, it requires costly per-scene optimization. Recent generalizable methods predict Gaussian representations from input images in a feed-forward manner, substantially improving reconstruction efficiency and generalization to unseen scenes~\cite{pixelsplat,mvsplat,freesplat,depthsplat,mvsgaussian}.

Most feed-forward 3D-GS methods decode one or more Gaussians from each pixel. Although this pixel-aligned formulation provides strong geometric anchoring, its representation size grows rapidly with image resolution and input-view number. Overlapping observations additionally produce many redundant Gaussians, increasing the storage and rendering costs of the final scene asset. Existing methods reduce pixel-aligned Gaussian redundancy through different designs. Geometry-aware fusion~\cite{freesplat,freesplat++} and view compression~\cite{zpressor} consolidate overlapping pixel-aligned predictions across views; voxel-aligned prediction~\cite{volsplat} predicts Gaussians in a shared 3D volume, whereas global scene encoders~\cite{c3g,globalsplat} compress multi-view observations into a compact set of scene tokens before decoding the Gaussian representation. The latter demonstrate that a fixed-capacity Gaussian set can be sufficient on some benchmarks, but its output size remains tied to the scene token capacity.

These developments raise a practical question: \emph{Can a single feed-forward model adapt its final Gaussian representation to different user-specified budgets?} Existing approaches generally fall into two categories. Fixed-capacity methods~\cite{c3g,globalsplat} tie the representation size to a particular token or decoder configuration, so substantially larger or more complex scenes may require retraining a higher-capacity model. Post-hoc compression methods~\cite{lightgaussian} reduce an already reconstructed Gaussian set, but introduce additional per-scene processing or optimization. In practice, deployment conditions may require the same model to operate under different representation budgets. Simply applying a budget-unaware Top-$K$ rule to a conventionally trained pixel-aligned model is suboptimal: it may over-allocate Gaussians to salient or high-opacity regions, leaving weakly textured areas under-covered. We therefore view budget-controlled Gaussian compaction as a joint problem of importance estimation, spatial coverage, and cross-view redundancy, rather than simple candidate ranking. These limitations motivate a feed-forward model whose final Gaussian representation can be adjusted at test time while retaining the geometric advantages of pixel-aligned reconstruction.

To this end, we propose \ours, a budget-controllable feed-forward 3D-GS framework that treats compact Gaussian allocation as a joint problem of importance estimation, spatial coverage, and cross-view redundancy. A single trained model supports varying context-view counts and user-specified test-time budgets without retraining or per-scene compression. Intra-Cross Mask Supervision provides structure-, coverage-, and redundancy-aware guidance for learning Gaussian importance, while Budget-Conditioned Feature Modulation adapts Gaussian prediction to the requested preservation ratio. Finally, Coverage-Aware Budget Selection combines local pre-selection, view-wise opacity ranking, and 3D redundancy suppression to construct the compact representation.

Our main contributions are summarized as follows:
\begin{itemize}
    \item We formulate budget-controllable feed-forward 3D-GS as a joint problem of importance estimation, spatial coverage, and cross-view redundancy, and propose a single pixel-aligned model that supports different context-view counts and user-specified test-time budgets without retraining or per-scene compression.

    \item We introduce Intra-Cross Mask Supervision to preserve informative structures, maintain sparse support in low-frequency regions, and reduce repeated cross-view supervision, together with Budget-Conditioned Feature Modulation that adapts Gaussian opacity and attributes to the requested preservation ratio.

    \item We introduce Coverage-Aware Budget Selection, which combines patch-wise pre-selection, view-wise global opacity ranking, and 3D redundancy suppression to produce compact yet spatially distributed Gaussian representations. Experiments on ScanNet and Replica demonstrate robust quality--size trade-offs across different input-view settings and test-time budgets.
\end{itemize}
\section{2. Related Work}
\label{sec:related}

\boldstart{Generalizable novel view synthesis.}
NeRF~\cite{nerf} and 3D Gaussian Splatting (3D-GS)~\cite{3dgs} achieve high-quality novel view synthesis but typically require per-scene optimization. Generalizable methods instead learn scene priors for feed-forward reconstruction. Early approaches predict image-conditioned radiance fields using pixel-aligned features, feature matching, or cost volumes~\cite{ibrnet,pixelnerf,mvsnerf,matchnerf,gefu}. Recent methods extend this paradigm to 3D-GS by directly predicting Gaussian geometry and attributes from input images~\cite{pixelsplat,mvsplat}. DepthSplat~\cite{depthsplat} and MonoSplat~\cite{monosplat} further incorporate pretrained monocular depth priors to improve geometry and generalization, but introduce additional dependence on external foundation models and feature adaptation. More importantly, these methods primarily target reconstruction quality and do not explicitly control the size of the final Gaussian representation.

\boldstart{Compact Gaussian representations.}
Pixel-aligned prediction provides strong geometric anchoring, but its Gaussian count scales with image resolution and input-view number, causing substantial cross-view redundancy. Geometry-aware fusion~\cite{freesplat,freesplat++} and view compression~\cite{zpressor} reduce repeated information, whereas voxel-aligned prediction~\cite{volsplat} and scene-level decoders~\cite{c3g,globalsplat} redesign the prediction space to produce more compact representations. However, their output size is determined by the underlying fusion or representation design rather than a user-specified test-time budget. Post-hoc methods such as LightGaussian~\cite{lightgaussian} compress an already reconstructed Gaussian set but require additional per-scene processing. EcoSplat~\cite{ecosplat} learns Gaussian importance and selects primitives through opacity ranking, but the final selection does not explicitly reason about spatial coverage or cross-view redundancy. In contrast, BudgetSplat treats controllable compaction as a joint importance--coverage--redundancy problem, introducing explicit low-frequency support, cross-view redundancy weighting, coverage-aware selection, and target-view alpha completion.

% -------------------------------------------------------
\begin{figure*}[!t]
    \centering
    \includegraphics[width=\textwidth]{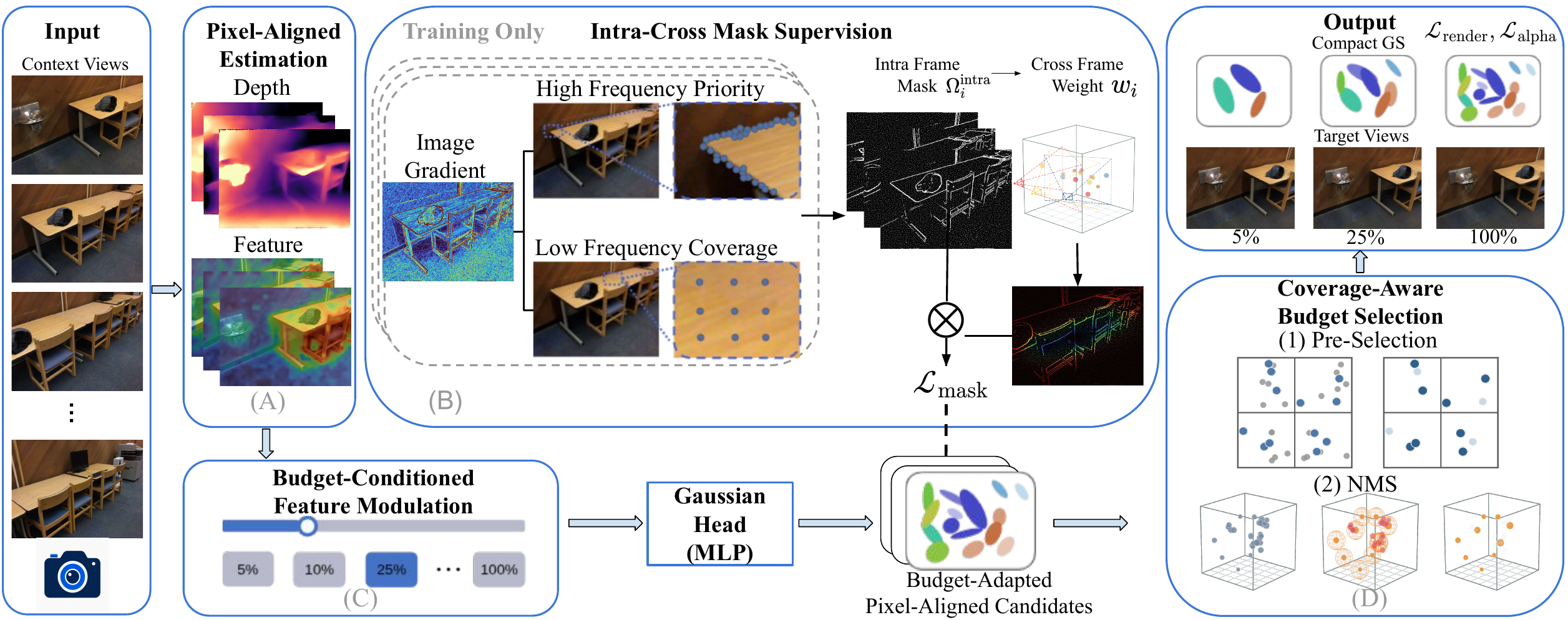}
    \caption{\textbf{Overview of \ours}. \textbf{(A)} Given $N$ posed context images, a pixel-aligned reconstruction backbone estimates dense geometry and base Gaussian features. \textbf{(B)} During training, Intra-Cross Mask Supervision constructs pseudo importance targets that preserve high-frequency structures, maintain sparse coverage in low-frequency regions, and down-weight redundant cross-view observations. \textbf{(C)} Budget-Conditioned Feature Modulation injects the user-specified preservation ratio $\rho$ into the base Gaussian features before parameter prediction, allowing the predicted opacity and Gaussian attributes to adapt to the requested budget. \textbf{(D)} Coverage-Aware Budget Selection combines patch-wise pre-selection, view-wise global opacity ranking, and 3D redundancy suppression to retain a compact and spatially distributed Gaussian representation. A single trained model supports arbitrary test-time preservation ratios $\rho\in(0,1]$ without retraining.}
    \label{fig:pipeline}
\end{figure*}
% -------------------------------------------------------

\section{3. Method}
\label{sec:method}

The overview of \ours is illustrated in Figure~\ref{fig:pipeline}. Given $N$ posed context images
$\mathcal{I}=\{{\bm I}^{i}\}_{i=1}^{N}$, where ${\bm I}^{i}\in\mathbb{R}^{H\times W\times 3}$, and their corresponding projection matrices $\mathcal{P}=\{{\bm P}^{i}\}_{i=1}^{N}$ with ${\bm P}^{i}=\mathbf{K}^{i}[\mathbf{R}^{i}\mid\mathbf{t}^{i}]$, our objective is to reconstruct a compact 3D Gaussian representation whose size can be controlled at inference time. Here, $\mathbf{K}^{i}$, $\mathbf{R}^{i}$, and $\mathbf{t}^{i}$ denote the camera intrinsic matrix, rotation, and translation, respectively.

Pixel-aligned feed-forward methods typically instantiate Gaussian candidates densely from the input images. As a result, the candidate count grows with both image resolution and the number of context views. We introduce a user-specified preservation ratio $\rho\in(0,1]$ to control the target fraction of candidates retained for rendering. Our goal is to learn the following budget-controllable mapping:
\begin{equation}
    f_{\bm{\theta}}:
    \left(
        \{({\bm I}^{i},{\bm P}^{i})\}_{i=1}^{N},
        \rho
    \right)
    \mapsto
    \{(\bm{\mu}_{j},\alpha_{j},
       \bm{\Sigma}_{j},\bm{c}_{j})\}_{j=1}^{M},
\end{equation}
where $M$ denotes the actual number of Gaussians in the final
representation.

The requested ratio defines a target budget
\begin{equation}
    K=\left\lfloor\rho NHW\right\rfloor.
\end{equation}
The subsequent geometric redundancy suppression may produce $M\leq K$, so $K$ is treated as an upper bound on the final representation size. An absolute upper-bound target budget $K_{\mathrm{abs}}$ can equivalently be specified through $\rho=K_{\mathrm{abs}}/(NHW)$.

The inference pipeline consists of three stages. First, the reconstruction backbone estimates dense pixel-aligned geometry and image features. Second, the requested ratio $\rho$ conditions the Gaussian prediction branch, allowing the predicted opacity and Gaussian attributes to adapt
to the target budget. Third, Coverage-Aware Budget Selection retains a spatially distributed subset under the requested budget. The pseudo masks, cross-view redundancy weights, and auxiliary losses introduced later are used only during training and incur no additional inference-time supervision.

\subsection{3.1 Pixel-aligned Gaussian Estimation}
\label{sec:pixel}
We adopt FreeSplat's feature extraction, cost-volume depth estimation, and Gaussian prediction backbone~\cite{freesplat}, while removing its native geometric fusion module. Given posed context images, the backbone extracts multi-scale features, constructs a multi-view cost volume, and estimates per-pixel depth. The depth is unprojected to obtain Gaussian centers $\bm{\mu}_i(\bm p)$ and corresponding base features
$\bm f_i(\bm p)$. As described in the next section, the requested budget is subsequently injected into this feature before the Gaussian parameters are predicted. The resulting candidates are then processed by our coverage-aware budget selection strategy to construct the final representation. We retain the architecture of the FreeSplat-based prediction backbone without modification, while optimizing it jointly with the proposed budget-controlled components.

\subsection{3.2 Budget-conditioned Gaussian Compaction}
\label{sec:budget}
The reconstruction backbone produces $NHW$ pixel-aligned
features and Gaussian centers. Given a requested preservation ratio $\rho$, we condition the features used for Gaussian prediction and select a subset for rasterization. This compaction remains part of the feed-forward inference pipeline and requires neither per-budget retraining nor per-scene optimization.

\boldstart{Budget-Conditioned Feature Modulation (BCFM).}
Changing only the selection budget would apply different truncation levels to budget-independent Gaussian predictions.
We therefore inject the requested ratio $\rho$ into the feature used by the Gaussian prediction heads. Specifically, $\rho$ is spatially broadcast and transformed by a lightweight convolutional embedding network into a budget feature $\bm E_i\in\mathbb R^{C\times H\times W}$, which is added to the base feature $\bm F_i\in\mathbb R^{C\times H\times W}$ before Gaussian parameter prediction:
\begin{equation}
    \left\{
    \tilde{\alpha}_{i},
    \tilde{\bm{\Sigma}}_{i},
    \tilde{\bm{c}}_{i}
    \right\}
    =
    \mathrm{MLP}
    \left(
        \bm F_i+\bm E_i
    \right).
\end{equation}
Here, $\tilde{\alpha}_{i}$, $\tilde{\bm{\Sigma}}_{i}$, and $\tilde{\bm{c}}_{i}$ denote the predicted opacity, covariance, and color, respectively. The activated opacity is used both for rendering and for candidate ranking. This conditioning allows a single model to adapt its Gaussian predictions to different test-time budgets.

\boldstart{Coverage-Aware Budget Selection (CABS).}
Directly ranking all candidates within a view may concentrate the retained Gaussians in salient regions and leave smooth or weakly textured areas under-covered. We therefore perform coverage-aware selection in three steps.
First, each input view is divided into local patches, from which an oversampled set of high-opacity candidates is retained to construct a spatially distributed candidate pool.
Second, all candidates retained from the patches of the same view are ranked jointly according to their predicted opacity.
We refer to this step as \emph{view-wise global opacity ranking}, since the ranking is performed across the candidate pool of one view rather than independently within each patch.
Finally, the ranked candidate pools from all views are jointly filtered by 3D non-maximum suppression to remove geometrically redundant candidates according to the requested budget.
The retained candidates form the compact Gaussian representation $\mathcal G_{\rho}$. Detailed budget allocation and selection procedures are provided in the supplementary material.

\subsection{3.3 Learning Gaussian Importance}
\label{sec:importance}
Since CABS ranks Gaussian candidates by their activated opacity, the opacity should reflect how useful each candidate is under specified budgets. However, ground-truth Gaussian importance labels are unavailable. We therefore introduce Intra-Cross Mask Supervision (ICMS), which consists of intra-frame structure-and-coverage mask construction, cross-frame redundancy weighting, and weighted opacity supervision. The first two components generate pseudo labels and supervision weights, which are jointly applied to the activated opacity through a weighted mask loss.

Let $\mathcal P=\{1,\ldots,H\}\times\{1,\ldots,W\}$ denote the image pixel grid. We use $i$ and $t$ to index context and target views, respectively, and $\bm p=(u,v)\in\mathcal P$ to denote a pixel location.

\boldstart{Intra-frame structure and coverage.}
For each context image $\bm I^i$, we compute an image-gradient
magnitude map:
\begin{equation}
    \bm g_i
    =
    \sqrt{
        \|\nabla_x \bm I^i\|_2^2
        +
        \|\nabla_y \bm I^i\|_2^2
    },
\end{equation}
where larger responses indicate detail-rich regions. To prevent the pseudo mask from allocating the entire preservation budget to high-frequency structures, we assign only a portion of the requested ratio to these regions and reserve the remainder for sparse low-frequency coverage. Specifically, we set $\rho_h=\gamma\rho$, with $\gamma=0.6$ in our experiments, and define
\begin{equation}
    \Omega^{\mathrm{high}}_i(\bm p)
    =
    \begin{cases}
        1,
        &
        \bm g_i(\bm p)>
        Q_{1-\rho_h}
        \left(
            \{\bm g_i(\bm q)\mid\bm q\in\mathcal P\}
        \right),
        \\
        0,
        &
        \text{otherwise},
    \end{cases}
\end{equation}
where $Q_{1-\rho_h}(\cdot)$ denotes the
$(1-\rho_h)$-quantile within the current view.

High-frequency selection alone may leave large smooth regions uncovered. We therefore uniformly sample a sparse subset $\mathcal L_i$ from the remaining pixels
\begin{equation}
    \mathcal U_i
    =
    \left\{
        \bm p\in\mathcal P
        \mid
        \Omega^{\mathrm{high}}_i(\bm p)=0
    \right\},
\end{equation}
and define
\begin{equation}
    \Omega^{\mathrm{low}}_i(\bm p)
    =
    \begin{cases}
        1, & \bm p\in\mathcal L_i,\\
        0, & \text{otherwise}.
    \end{cases}
\end{equation}
Since the high- and low-frequency sets are disjoint, the intra-frame pseudo mask is
\begin{equation}
    \Omega^{\mathrm{intra}}_i
    =
    \Omega^{\mathrm{high}}_i
    +
    \Omega^{\mathrm{low}}_i.
\end{equation}
Pixels selected from either branch are labeled as positive, while all remaining pixels are labeled as negative. This construction favors detail-rich regions while retaining sparse support over weakly textured areas.

\boldstart{Cross-frame redundancy weighting.}
Positive candidates from different context views may observe the same 3D region. We keep their binary pseudo labels unchanged, but reduce the repeated positive supervision. Each positive pixel is first unprojected into the global 3D space:
\begin{equation}
    \bm x_{i,\bm p}
    =
    \Pi^{-1}
    \left(
        \bm p,
        d_i(\bm p),
        \bm P^i
    \right).
\end{equation}

For high-frequency candidates, we first retain one representative from each occupied coarse voxel independently within every context view. The retained representatives are then grouped across views using a finer voxel grid. Let $\mathcal S_{\mathrm{high}}$ denote the retained representatives, and let $\mathcal V_i(\bm p)$ denote the fine-voxel group containing candidate $(i,\bm p)$. We define the positive supervision weight as
\begin{equation}
    w_i^{+}(\bm p)
    =
    \begin{cases}
        \lambda_{\mathrm{low}},
        & \Omega_i^{\mathrm{low}}(\bm p)=1,\\[3pt]
        \dfrac{1}{|\mathcal V_i(\bm p)|},
        & (i,\bm p)\in\mathcal S_{\mathrm{high}},\\[5pt]
        0,
        & \text{otherwise}.
    \end{cases}
\end{equation}
Therefore, low-frequency support receives a fixed weaker weight, while the representatives of each repeated 3D region share a total positive weight of one.

\boldstart{ICMS objective.}
Positive pixels use the weights defined above, whereas negative pixels retain unit weight:
\begin{equation}
    \bar w_i(\bm p)
    =
    \begin{cases}
        w_i^{+}(\bm p),
        & \Omega_i^{\mathrm{intra}}(\bm p)=1,\\
        1,
        & \Omega_i^{\mathrm{intra}}(\bm p)=0.
    \end{cases}
\end{equation}
We supervise the activated opacity using weighted binary cross-entropy:
\begin{equation}
    \mathcal L_{\mathrm{mask}}
    =
    \frac{1}{NHW}
    \sum_{i=1}^{N}
    \sum_{\bm p\in\mathcal P}
    \bar w_i(\bm p)\,
    \mathcal L_{\mathrm{BCE}}
    \left(
        \Omega_i^{\mathrm{intra}}(\bm p),
        \tilde{\alpha}_i(\bm p)
    \right).
\end{equation}
Thus, cross-frame redundancy weighting suppresses repeated positive supervision without changing the standard negative opacity supervision.

\boldstart{Alpha completion.}
The weighted mask loss provides local supervision for Gaussian importance, but does not directly guarantee complete coverage in the target views, especially under aggressive budgets. We therefore render the compact Gaussian representation $\mathcal G_{\rho}$ and penalize missing alpha support:
\begin{equation}
    \mathcal L_{\mathrm{alpha}}
    =
    \frac{1}{N^{\mathrm{tgt}}HW}
    \sum_{t=1}^{N^{\mathrm{tgt}}}
    \sum_{\bm p\in\mathcal P}
    \left|
        1-\bm A^{\mathrm{tgt}}_t(\bm p)
    \right|,
\end{equation}
where $\bm A^{\mathrm{tgt}}_t(\bm p)$ denotes the rendered alpha value at pixel $\bm p$ in target view $t$. This auxiliary loss complements the local importance supervision by encouraging the selected Gaussians to maintain complete target-view coverage.

\subsection{3.4 Training Strategy and Objective}
\label{sec:loss}

We train \ours using ground-truth target RGB images in two stages. The first stage learns dense pixel-aligned reconstruction without budget-controlled selection. Following prior works~\cite{pixelsplat,mvsplat,freesplat,depthsplat}, we optimize the rendered target image using
\begin{equation}
    \mathcal{L}_{\mathrm{render}}
    =
    \lambda_{\mathrm{MSE}}
    \mathcal{L}_{\mathrm{MSE}}
    +
    \lambda_{\mathrm{LPIPS}}
    \mathcal{L}_{\mathrm{LPIPS}}.
\end{equation}
This stage establishes a reliable dense reconstruction before budget-controlled compaction is introduced.

After the dense prediction converges, we activate the budget-controlled branch and continue training with top-$K$ rendering. During this stage, $\rho$ is randomly sampled and is injected into the prediction branch through BCFM. Its sampling range is gradually shifted toward lower preservation ratios. The compact representation produced under the sampled budget is rasterized, and the model is jointly optimized using the reconstruction, mask supervision, and alpha completion objectives:
\begin{equation}
    \mathcal{L}
    =
    \mathcal{L}_{\mathrm{render}}
    +
    \lambda_{\mathrm{mask}}
    \mathcal{L}_{\mathrm{mask}}
    +
    \lambda_{\mathrm{alpha}}
    \mathcal{L}_{\mathrm{alpha}} .
\end{equation}
Through this two-stage procedure, the model first learns reliable dense reconstruction and then learns to adapt its Gaussian prediction and selection to the requested budget while preserving spatial coverage.

% -------------------------------------------------------
\begin{figure*}[!t]
    \centering
    \includegraphics[width=\textwidth]{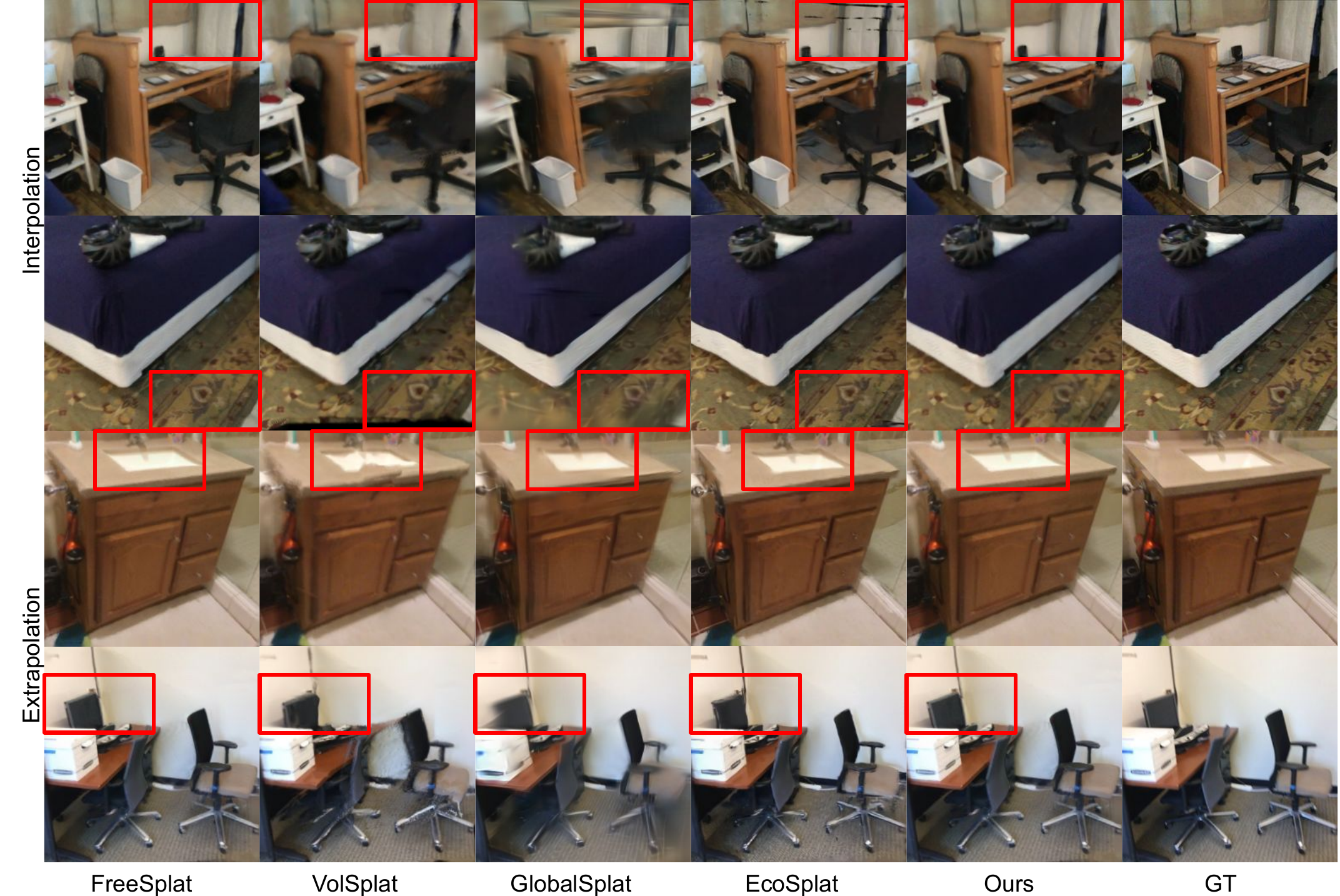}
    \caption{\textbf{Qualitative results on ScanNet~\cite{scannet} with 10 context views.} Ours is rendered using only $25\%$ of the Gaussian budget.}
    \label{fig:scannet_10_views_prune}
\end{figure*}
% -------------------------------------------------------

\section{4. Experiments}
\label{sec:exp}

\subsection{4.1 Experimental Setup}
\label{sec:exp_setup}

\boldstart{Datasets.}
Following FreeSplat~\cite{freesplat}, we train on the same
ScanNet~\cite{scannet} split with 100 training scenes and 8 test scenes, and evaluate zero-shot generalization on 8 Replica~\cite{replica} scenes.
We report standard image quality metrics, including pixel-level PSNR, patch-level SSIM~\cite{ssim}, and feature-level LPIPS~\cite{lpips}.

\boldstart{Implementation details.}
We train using Adam~\cite{adam} with a learning rate of $10^{-4}$ and a batch size of $1$ on a single RTX 4090 GPU. Unless otherwise specified, images are resized to $384\times512$. We first train the dense reconstruction model for 300K iterations. Starting from this checkpoint, we train the budget-controlled stage for another 300K iterations. The final model therefore undergoes 600K iterations in total. We set
$\lambda_{\mathrm{MSE}}=1.0$,
$\lambda_{\mathrm{LPIPS}}=0.05$,
$\lambda_{\mathrm{mask}}=0.1$, and
$\lambda_{\mathrm{alpha}}=1.0$.
Detailed training schedules are provided in the supplementary material.

\boldstart{Baselines.}
We group compared methods into native-capacity (pixelSplat~\cite{pixelsplat}, MVSplat~\cite{mvsplat},
DepthSplat~\cite{depthsplat}, FreeSplat~\cite{freesplat},
ZPressor~\cite{zpressor}, VolSplat~\cite{volsplat}, and
GlobalSplat~\cite{globalsplat}) and budgeted-compaction
(EcoSplat~\cite{ecosplat}) methods. Unless otherwise noted, we use official checkpoints and configurations under the same view splits. We include the published 2- and 3-view results of FreeSplat++~\cite{freesplat++}, but omit it from the 10-view comparison because its implementation is unavailable for reproducing our $256\times256$ protocol. BudgetSplat$_{X\%}$ denotes the same trained model evaluated with $\rho=X/100$, and \#GS is the number of rendered Gaussians. \textbf{Bold} and \underline{underlined} values indicate the best and second-best results within each group.

%%%%%%%%%%%%%%%%%%%%%%%%%%%%%
% NVS on ScanNet
\begin{table*}[!t]
    \centering
    \small
    
    \setlength{\tabcolsep}{1.0mm}{
    \begin{tabular}{llcccccccc}
        \toprule
        \multirow{2}{*}{Type}
        & \multirow{2}{*}{Method}
        & \multicolumn{4}{c}{2 Views}
        & \multicolumn{4}{c}{3 Views} \\
        \cmidrule(lr){3-6}
        \cmidrule(lr){7-10}
        &
        & PSNR$\uparrow$
        & SSIM$\uparrow$
        & LPIPS$\downarrow$
        & \#GS(k)$\downarrow$
        & PSNR$\uparrow$
        & SSIM$\uparrow$
        & LPIPS$\downarrow$
        & \#GS(k)$\downarrow$ \\
        \midrule

        \multirow{7}{*}{\shortstack[c]{Native\\capacity}}
        & pixelSplat~\cite{pixelsplat}
        & 24.50
        & 0.776
        & 0.288
        & 1180
        & 23.44
        & \underline{0.782}
        & 0.270
        & 1769 \\

        & MVSplat~\cite{mvsplat}
        & 24.02
        & 0.771
        & \underline{0.268}
        & 393
        & 22.26
        & 0.744
        & 0.328
        & 590 \\

        & DepthSplat~\cite{depthsplat}
        & 24.59
        & 0.783
        & 0.276
        & 393
        & 23.10
        & 0.761
        & 0.327
        & 590 \\

        & FreeSplat~\cite{freesplat}
        & \underline{28.08}
        & \underline{0.837}
        & \textbf{0.211}
        & 278
        & \textbf{27.45}
        & \textbf{0.829}
        & \textbf{0.222}
        & 382 \\

        & FreeSplat++$^\dagger$~\cite{freesplat++}
        & \textbf{28.19}
        & \textbf{0.838}
        & \underline{0.212}
        & \underline{68}
        & \textbf{27.45}
        & \textbf{0.829}
        & \underline{0.223}
        & \underline{89} \\

        & ZPressor~\cite{zpressor}
        & 18.32
        & 0.694
        & 0.369
        & 393
        & 21.65
        & 0.746
        & 0.317
        & 590 \\

        & VolSplat~\cite{volsplat}
        & 23.40
        & \underline{0.783}
        & 0.365
        & 393
        & 22.99
        & 0.775
        & 0.372
        & 590 \\

        & GlobalSplat~\cite{globalsplat}
        & 21.81
        & 0.725
        & 0.419
        & \textbf{33}
        & \underline{23.49}
        & 0.773
        & 0.383
        & \textbf{33} \\

        \midrule

        \multirow{4}{*}{\shortstack[c]{Budgeted\\compaction}}
        & EcoSplat$_{100\%}$~\cite{ecosplat}
        & 21.89
        & 0.747
        & 0.352
        & 393
        & 22.15
        & 0.752
        & 0.358
        & 590 \\

        & EcoSplat$_{25\%}$~\cite{ecosplat}
        & 19.97
        & 0.672
        & 0.426
        & \textbf{98}
        & 20.35
        & 0.703
        & 0.420
        & \textbf{147} \\

        % & F4Splat~\cite{f4splat}
        % & 19.36
        % & 0.671
        % & 0.434
        % & \textbf{98}
        % & 19.59
        % & 0.689
        % & 0.444
        % & \textbf{147} \\

        & \textbf{BudgetSplat$_{100\%}$}
        & \textbf{28.11}
        & \underline{0.839}
        & \textbf{0.228}
        & \underline{382}
        & \textbf{27.13}
        & \underline{0.825}
        & \textbf{0.245}
        & \underline{565} \\

        & \textbf{BudgetSplat$_{25\%}$}
        & \underline{27.84}
        & \textbf{0.840}
        & \underline{0.250}
        & \textbf{98}
        & \underline{27.07}
        & \textbf{0.828}
        & \underline{0.264}
        & \textbf{147} \\

        % & \ours$_{5\%}$
        % & \underline{26.41}
        % & \textbf{0.813}
        % & \underline{0.355}
        % & \textbf{19}
        % & \underline{26.14}
        % & \textbf{0.814}
        % & \underline{0.349}
        % & \textbf{29} \\

        \bottomrule
    \end{tabular}

    \caption{
    \textbf{Generalizable NVS interpolation results on ScanNet~\cite{scannet} with 2 and 3 context views.}
    The input resolution is $384\times512$. $\dagger$ denotes results reported in the original paper.}
    \label{tab:scannet_generalization}
    }

\end{table*}
%%%%%%%%%%%%%%%%%%%%%%%%%%%%%

%%%%%%%%%%%%%%%%%%%%%%%%%%%%%
% Long on ScanNet
\begin{table*}[!t]
    \centering
    \small

    \setlength{\tabcolsep}{1.0mm}{
    \begin{tabular}{llcccccccc}
        \toprule
        \multirow{2}{*}{Type}
        & \multirow{2}{*}{Method}
        & \multirow{2}{*}{\#GS(k)$\downarrow$}
        & \multirow{2}{*}{\shortstack{GS Size\\(MiB)$\downarrow$}}
        & \multicolumn{3}{c}{View Interpolation}
        & \multicolumn{3}{c}{View Extrapolation} \\
        \cmidrule(lr){5-7}
        \cmidrule(lr){8-10}
        &
        &
        &
        & PSNR$\uparrow$
        & SSIM$\uparrow$
        & LPIPS$\downarrow$
        & PSNR$\uparrow$
        & SSIM$\uparrow$
        & LPIPS$\downarrow$ \\
        \midrule

        \multirow{7}{*}{\shortstack[c]{Native\\capacity}}
        & pixelSplat~\cite{pixelsplat}
        & 1966
        & 442.5
        & 20.31
        & 0.713
        & 0.375
        & 18.93
        & 0.680
        & 0.410 \\

        & MVSplat~\cite{mvsplat}
        & 655
        & 147.4
        & 19.25
        & 0.661
        & 0.379
        & 19.00
        & 0.663
        & 0.382 \\

        & DepthSplat~\cite{depthsplat}
        & 655
        & 147.4
        & 21.76
        & 0.735
        & \underline{0.297}
        & 21.10
        & 0.720
        & 0.311 \\

        & FreeSplat~\cite{freesplat}
        & \underline{294}
        & \underline{66.2}
        & \textbf{25.03}
        & \textbf{0.800}
        & \textbf{0.223}
        & \textbf{23.96}
        & \textbf{0.780}
        & \textbf{0.250} \\

        & ZPressor~\cite{zpressor}
        & 393
        & 88.5
        & 21.59
        & 0.746
        & 0.317
        & 21.43
        & 0.730
        & \underline{0.289} \\

        & VolSplat~\cite{volsplat}
        & 655
        & 147.4
        & 22.49
        & \underline{0.769}
        & 0.325
        & \underline{21.82}
        & \underline{0.754}
        & 0.341 \\

        & GlobalSplat~\cite{globalsplat}
        & \textbf{33}
        & \textbf{7.4}
        & \underline{23.13}
        & 0.761
        & 0.338
        & \underline{21.82}
        & 0.721
        & 0.381 \\

        \midrule

        \multirow{5}{*}{\shortstack[c]{Budgeted\\compaction}}
        & EcoSplat$_{100\%}$~\cite{ecosplat}
        & 655
        & 147.4
        & 21.97
        & 0.761
        & 0.306
        & 22.08
        & 0.753
        & \underline{0.311} \\

        & EcoSplat$_{25\%}$~\cite{ecosplat}
        & \underline{163}
        & \underline{36.7}
        & 21.87
        & 0.766
        & 0.315
        & 21.81
        & 0.751
        & 0.330 \\

        & EcoSplat$_{5\%}$~\cite{ecosplat}
        & \textbf{33}
        & \textbf{7.4}
        & 21.43
        & 0.751
        & 0.354
        & 21.31
        & 0.735
        & 0.370 \\

        % & F4Splat~\cite{f4splat}
        % & \underline{163}
        % & \underline{36.7}
        % & 22.28
        % & 0.761
        % & 0.325
        % & 22.07
        % & 0.756
        % & \underline{0.327} \\

        & \textbf{BudgetSplat$_{100\%}$}
        & 653
        & 147.0
        & \underline{24.90}
        & \underline{0.790}
        & \textbf{0.254}
        & \textbf{23.70}
        & \underline{0.771}
        & \textbf{0.280} \\

        & \textbf{BudgetSplat$_{25\%}$}
        & \underline{163}
        & \underline{36.7}
        & \textbf{24.91}
        & \textbf{0.793}
        & \underline{0.295}
        & \underline{23.66}
        & \textbf{0.775}
        & 0.318 \\

        & \textbf{BudgetSplat$_{5\%}$}
        & \textbf{33}
        & \textbf{7.4}
        & 24.68
        & 0.789
        & 0.368
        & 23.35
        & \underline{0.771}
        & 0.391 \\

        \bottomrule
    \end{tabular}

    \caption{
    \textbf{Generalizable NVS interpolation and extrapolation results on ScanNet~\cite{scannet} with 10 context views.} The input resolution is $256\times256$.}
    \label{tab:scannet_long}
    }
    
\end{table*}

\subsection{4.2 Comparisons with Existing Methods}
\label{sec:exp_comparisons}

\boldstart{Results on ScanNet.}
We evaluate our method on ScanNet~\cite{scannet} using 2, 3, and 10 context views. All settings use the same trained model, which also supports arbitrary preservation ratios at test time without retraining. Note that \ours$_{100\%}$ contains more Gaussians than native FreeSplat~\cite{freesplat} because our model operates on the dense pixel-aligned candidate set and does not use its native geometric fusion. The lower-budget variants therefore reflect compaction by our selector rather than additional pruning of FreeSplat's already fused output.

Figure~\ref{fig:scannet_10_views_prune} compares \ours with recent feed-forward 3D-GS methods. While competing methods exhibit incomplete coverage, structural discontinuities, or local over-smoothing, \ours better preserves scene completeness and fine details under a compact budget. The supplementary material further visualizes both the selected Gaussian centers and the corresponding renderings across different preservation ratios.

As shown in Table~\ref{tab:scannet_generalization}, \ours achieves the best rendering quality among the evaluated budgeted-compaction methods. At the $25\%$ budget, it matches EcoSplat~\cite{ecosplat}'s Gaussian counts, while substantially improving all rendering metrics. Compared with FreeSplat~\cite{freesplat}, it uses over $60\%$ fewer Gaussians with PSNR differences of only $0.24$ and $0.38$ dB.

Table~\ref{tab:scannet_long} further reports interpolation and extrapolation results with 10 context views. All methods are evaluated at $256\times256$ because several pixel-aligned baselines run out of memory at $384\times512$ on an RTX 4090. BudgetSplat$_{25\%}$ achieves the best PSNR and SSIM among the budgeted-compaction methods and remains close to FreeSplat~\cite{freesplat} in PSNR and SSIM, while using approximately $45\%$ fewer Gaussians. Even at the aggressive $5\%$ budget, \ours uses only $33$k Gaussians, with PSNR decreasing by $0.22$ dB for interpolation and $0.35$ dB for extrapolation relative to BudgetSplat$_{100\%}$.

%%%%%%%%%%%%%%%%%%%%%%%%%%%%%
% Replica
\begin{table*}[!t]
    \centering
    \small
    
    \setlength{\tabcolsep}{1.0mm}
    {
    \begin{tabular}{llcccccccc}
        \toprule
        \multirow{2}{*}{Type}
        & \multirow{2}{*}{Method}
        & \multicolumn{4}{c}{3 Views}
        & \multicolumn{4}{c}{10 Views} \\
        \cmidrule(lr){3-6}
        \cmidrule(lr){7-10}
        &
        & PSNR$\uparrow$
        & SSIM$\uparrow$
        & LPIPS$\downarrow$
        & \#GS(k)$\downarrow$
        & PSNR$\uparrow$
        & SSIM$\uparrow$
        & LPIPS$\downarrow$
        & \#GS(k)$\downarrow$ \\
        \midrule

        \multirow{7}{*}{\shortstack[c]{Native\\capacity}}
        & pixelSplat~\cite{pixelsplat}
        & 25.88
        & 0.838
        & 0.214
        & 1769
        & 21.03
        & 0.745
        & 0.344
        & 1966 \\

        & MVSplat~\cite{mvsplat}
        & 23.81
        & 0.812
        & 0.221
        & 590
        & 19.11
        & 0.696
        & 0.348
        & 655 \\

        & DepthSplat~\cite{depthsplat}
        & 25.73
        & 0.844
        & \textbf{0.169}
        & 590
        & 20.29
        & 0.733
        & \underline{0.275}
        & 655 \\

        & FreeSplat~\cite{freesplat}
        & \textbf{26.98}
        & \underline{0.848}
        & \underline{0.171}
        & \underline{423}
        & 21.24
        & 0.752
        & 0.289
        & 448 \\

        & ZPressor~\cite{zpressor}
        & 23.93
        & 0.816
        & 0.197
        & 590
        & 21.23
        & 0.750
        & \textbf{0.258}
        & \underline{393} \\

        & VolSplat~\cite{volsplat}
        & \underline{26.67}
        & \textbf{0.861}
        & 0.206
        & 590
        & \underline{21.70}
        & \textbf{0.774}
        & 0.290
        & 655 \\

        & GlobalSplat~\cite{globalsplat}
        & 25.99
        & 0.823
        & 0.286
        & \textbf{33}
        & \textbf{21.88}
        & \underline{0.755}
        & 0.305
        & \textbf{33} \\

        \midrule

        \multirow{4}{*}{\shortstack[c]{Budgeted\\compaction}}
        & EcoSplat$_{100\%}$~\cite{ecosplat}
        & 23.85
        & \underline{0.793}
        & 0.244
        & 590
        & 20.38
        & 0.715
        & \underline{0.305}
        & 655 \\

        & EcoSplat$_{25\%}$~\cite{ecosplat}
        & 22.53
        & 0.776
        & 0.299
        & \textbf{147}
        & 20.64
        & 0.730
        & 0.316
        & \textbf{163} \\

        % & F4Splat~\cite{f4splat}
        % & 20.98
        % & 0.744
        % & 0.350
        % & \textbf{147}
        % & \textbf{23.79}
        % & \textbf{0.811}
        % & \textbf{0.245}
        % & \textbf{163} \\

        & \textbf{BudgetSplat$_{100\%}$}
        & \textbf{26.38}
        & \textbf{0.841}
        & \textbf{0.211}
        & \underline{580}
        & \underline{21.08}
        & \underline{0.744}
        & \textbf{0.297}
        & \underline{631} \\

        & \textbf{BudgetSplat$_{25\%}$}
        & \underline{26.35}
        & \textbf{0.841}
        & \underline{0.235}
        & \textbf{147}
        & \textbf{21.14}
        & \textbf{0.747}
        & 0.313
        & \textbf{163} \\

        \bottomrule
    \end{tabular}

    \caption{
    \textbf{Cross-dataset generalization on Replica~\cite{replica}.}
    The input resolution is $384\times512$ for the 3-view setting and $256\times256$ for the 10-view setting.}
    \label{tab:replica_generalization}
    }

\end{table*}

\boldstart{Representation compactness.}
Because recent methods use different architectures and inference pipelines, we compare the output-side cost of the final Gaussian representation. Table~\ref{tab:scannet_long} also reports the Gaussian count and raw storage required for rendering. GlobalSplat~\cite{globalsplat} produces the smallest native representation, while BudgetSplat$_{5\%}$ achieves higher PSNR and SSIM with a comparable budget of about $33$k Gaussians. At the $25\%$ budget, \ours also clearly outperforms EcoSplat~\cite{ecosplat} using the same number of Gaussians. These results show that the preservation ratio enables a controllable trade-off between representation size and rendering quality.

%%%%%%%%%%%%%%%%%%%%%%%%%%%%
% Ablation on ScanNet
\begin{table}[!t]
    \centering
    \small
    
    \setlength{\tabcolsep}{1.0mm}
    \begin{tabular}{@{}llcccc@{}}
        \toprule
        Method
        & Components
        & PSNR$\uparrow$
        & SSIM$\uparrow$
        & LPIPS$\downarrow$
        & \#GS(k)$\downarrow$ \\
        \midrule

        % FreeSplat
        % & 
        % & 25.94
        % & 0.792
        % & 0.247
        % & 384 \\

        % \midrule

        Baseline
        & 
        & 13.02
        & 0.386
        & 0.561
        & 147 \\

        No.1
        & I
        & 24.80
        & 0.743
        & 0.336
        & 147 \\

        No.2
        & I + A
        & 26.14
        & 0.810
        & 0.311
        & 147 \\

        No.3
        & I + C + A
        & 26.38
        & 0.813
        & 0.310
        & 147 \\

        No.4
        & I + A + S
        & 26.75
        & 0.821
        & 0.279
        & 147 \\

        Full
        & I + C + A + S
        & \textbf{26.77}
        & \textbf{0.823}
        & \textbf{0.275}
        & 147 \\

        \bottomrule
    \end{tabular}%
    
    \caption{\textbf{Ablation on ScanNet~\cite{scannet}.}
    I, C, A, and S denote intra-frame supervision,
    cross-frame weighting, alpha completion, and
    coverage-aware budget selection, respectively.
    Results are reported using 3 context views.}
    \label{tab:ablation}

\end{table}
%%%%%%%%%%%%%%%%%%%%%%%%%%%%

\boldstart{Zero-shot generalization on Replica.}
We further evaluate cross-dataset generalization by applying the models to Replica~\cite{replica} without finetuning. As shown in Table~\ref{tab:replica_generalization}, we distinguish methods evaluated at their native output capacities from methods that explicitly perform budgeted Gaussian compaction. Among the budgeted-compaction methods, \ours achieves the strongest rendering quality. The results also show that absolute cross-dataset performance remains strongly influenced by the underlying reconstruction architecture and scene representation, as evidenced by the strong performance of VolSplat~\cite{volsplat} and GlobalSplat~\cite{globalsplat} in different settings.

More importantly, \ours maintains stable rendering quality when the budget is reduced from $100\%$ to $25\%$. These results indicate that the learned budget-conditioned selection generalizes to unseen scenes and removes approximately $74\%$ of the Gaussians with only limited additional degradation relative to the full representation.

\subsection{4.3 Ablations} 
\label{sec:exp_ablation}

We conduct ablations on ScanNet~\cite{scannet}. 
Starting from the 300K dense-stage checkpoint we trained, all variants are further trained for 100K iterations for budget compaction. Table~\ref{tab:ablation} evaluates the contributions of the supervision and selection components while keeping BCFM fixed. Additional ablations are provided in the supplementary material.

Our baseline applies plain view-wise opacity Top-K to retain 25\% of the dense pixel-aligned Gaussians, without importance supervision or coverage-aware selection, resulting in severe performance degradation.
Adding intra-frame structure-and-coverage supervision in No.1 substantially improves all rendering metrics, showing that the pseudo masks provide an effective signal for learning Gaussian importance.
No.2 further introduces alpha completion, which improves all metrics. This indicates that local mask supervision alone does not guarantee complete target-view coverage, particularly under aggressive compression.
No.3 adds cross-frame redundancy weighting to No.2. The resulting improvement shows that reducing the supervision assigned to repeated observations encourages a less redundant allocation of Gaussians across views.
No.4 adds coverage-aware budget selection to No.2. Its larger improvement demonstrates that spatially distributed candidate selection helps prevent the retained Gaussians from collapsing into highly textured regions.
The full model achieves the best overall performance, confirming that the training-time importance guidance and inference-time selection strategy are complementary.
\section{5. Conclusion}
\label{sec:conclusion}
We present \ours, a budget-controllable feed-forward 3D-GS. Instead of keeping all pixel-aligned Gaussians, \ours learns to compact them according to a user-specified budget ratio.
\ours can flexibly adjust the number of Gaussians while preserving rendering quality, without per-budget retraining or a separate offline compression stage. Experiments show that \ours provides an effective trade-off between visual quality, representation storage, rendering memory, and rendering efficiency.

\boldstart{Limitations and discussions.}
\ours relies on image gradients and estimated geometry for pseudo supervision, so degraded image quality or depth errors may affect selection under very compact budgets. Moreover, while \ours reduces the final representation and rendering costs, it retains the dense pixel-aligned prediction backbone. Future work will explore more robust supervision and budget-aware candidate generation.
%%%%%%%%% BODY TEXT END

% -------------------------------------------------------
{
    \small
    \bibliographystyle{ieeenat_fullname}
    \bibliography{main}
}

% %%%%%%%%% APPENDIX START
% \appendix
% \input{sec/6_supp}
% %%%%%%%%% APPENDIX END

% Check whether the conference requires a reproducibility checklist to be included in the paper.
% If so, you can uncomment the following line and ajust the path to include it.
% \input{ReproducibilityChecklist.tex}

\end{document}